\documentclass[sigconf]{acmart}
\usepackage{multirow}
\usepackage{multicol}
\usepackage{graphicx}
\usepackage{subfigure}
\usepackage[utf8]{inputenc}
\usepackage[russian, english]{babel}
\babeltags{ru = russian}

\renewcommand\footnotetextcopyrightpermission[1]{} 

\def\BibTeX{{\rm B\kern-.05em{\sc i\kern-.025em b}\kern-.08emT\kern-.1667em\lower.7ex\hbox{E}\kern-.125emX}}
\AtBeginDocument{%
  \providecommand\BibTeX{{%
    \normalfont B\kern-0.5em{\scshape i\kern-0.25em b}\kern-0.8em\TeX}}}

\setcopyright{rightsretained}
\acmConference[SIGIR eCom'20]{ACM SIGIR Workshop on eCommerce}{July 30, 2020}{Virtual Event, China}
\acmYear{2020}
\copyrightyear{2020}
\makeatletter
\renewcommand\@formatdoi[1]{\ignorespaces}
\makeatother
\acmISBN{}

\begin{document}

\title{Constraint Translation Candidates: A Bridge between Neural \\ Query Translation and Cross-lingual Information Retrieval}

\author{Tianchi Bi \qquad Liang Yao \qquad Baosong Yang \qquad Haibo Zhang}
\authornote{Corresponding author.}
\author{ Weihua Luo \qquad Boxing Chen} 
\affiliation{
\institution{Alibaba Group}
\city{Hangzhou, China}
}
\email{ {tianchi.btc, yaoliang.yl, yangbaosong.ybs, zhanhui.zhb, weihua.luowh, boxing.cbx}@alibaba-inc.com }  

\renewcommand{\shortauthors}{Bi et al.}

\begin{abstract}
Query translation (QT) is a key component in cross-lingual information retrieval system (CLIR). With the help of deep learning, neural machine translation (NMT) has shown promising results on various tasks. 
However, NMT is generally trained with large-scale out-of-domain data rather than in-domain query translation pairs. 
Besides, the translation model lacks a mechanism  at the inference time to guarantee the generated words to match the search index. 
The two shortages of QT result in readable texts for human but inadequate candidates for the downstream retrieval task. 
In this paper, we propose a novel approach to alleviate these problems by limiting the open target vocabulary search space of QT to a set of important words mined from search index database. 
The constraint translation candidates are employed at both of training and inference time, thus guiding the translation model to learn and generate well performing target queries.
The proposed methods are exploited and examined in a real-word CLIR system--Aliexpress e-Commerce search engine.\footnote{We have already exploited NMT into the real-world CLIR system, e.g. Aliexpress. Readers can check their cases on \url{https://aliexpress.com.}} Experimental results demonstrate that our approach yields better performance on both translation quality and retrieval accuracy than the strong NMT baseline. 

\end{abstract}




\maketitle

\section{Introduction}
Cross-lingual information retrieval (CLIR) can have separate query translation (QT), information retrieval (IR), as well as machine-learned ranking stages. Among them, QT stage takes a multilingual user query as input and returns the translation candidates in language of search index for the downstream retrieval.
To this end, QT plays a key role and its output significantly affects the retrieval results\cite{wu2010study,yao2020exploiting,yao2020domain}. 
In order to improve the translation quality, many efforts have been made based on techniques in machine translation community, e.g. bilingual dictionaries and statistical machine translation \cite{koehn2009statistical,och2002discriminative}. 
Recently, neural machine translation (NMT) has shown their superiority in a variety of translation tasks~\cite{ott2018scaling,hassan2018achieving}. Several studies begin to explore the feasibility and improvements of NMT for QT task \cite {sarwar2019multi,sharma2019refined}.

Nevertheless, taking the translation quality as the primary  optimization objective for neural query translation may fail to further improve the retrieval performance. Recent studies have pointed out that there seems no strong correlation between translation and retrieval qualities \cite{rubino2020effect,yarmohammadi2019robust}. For example, Fuji et al., \cite{Fujii09} empirically investigated this problem, and found the system with the highest human evaluation score in terms of translation, gained the relatively worse retrieval quality. Yarmohammadi et al., \cite{yarmohammadi2019robust} also noticed that NMT even has much higher missed detection rate compared to its SMT counterpart, despite its high translation accuracy. 

We attribute the mismatch between NMT and CLIR to two reasons. Firstly, a well-performed NMT model depends on extensive language resources~\cite{popel2018training,ott2018scaling,wan2020AAAI,Yang2020improve}, while the lack of in-domain query pairs leads existing neural query translation models to be trained using general domain data. This makes a well-trained NMT model fail since the vocabulary and style mismatch between the translated query and terms in search index. On the other hand,  the translation model lacks a mechanism to guarantee the produced words to be highly likely in search index at the inference time, resulting in readable texts for human but unaware candidates for the downstream retrieval task~\cite{zhou2012translation,sarwar2019multi}.


In this paper, we propose to alleviate the mentioned problems by restricting the generated target terms of NMT to  constraint candidates of which can be aware by information retrieval system. 
Since the target search index is built pursuant to the probability distribution of terms in documents, a natural way is to transfer the translation to those target candidates being likely to appear in the retrieval entries. 
Specifically, given a source query, we mined its constrained target terms according to the distribution of words in the entries clicked by users.
The large-scale cross-lingual clickthrough data on a real-world CLIR engine makes the proposed mining approach feasible and low cost. 

We exploit these constraint translation candidates at both of the training and predicting time. For the former, the candidates are served as the smoothed labels during the loss estimation. The NMT model is therefore guided to learn the distribution of search index. For the latter, we limit the output words to the collected candidates with the help of Weighted Softmax. In this way, the search-aware terms offer a bridge between neural query translation and 
information retrieval.


We build our model upon an advanced neural machine translation architecture--Transformer~\cite{vaswani2017attention,devlin2019bert} and evaluate the effectiveness of the proposed approach in a real-word e-Commerce search engine--Aliexpress. Experimental results demonstrate that the proposed method is able to improve the retrieval accuracy, at the same time, maintain the translation quality. The qualitative analysis confirms that our method exactly raises the ability of NMT to generates more suitable target queries for the scenario of e-Commerce search. 

\section{Background}


\subsection{Neural Machine Translation} \label{section:mt}


Neural machine translation (NMT) \cite{bahdanau2014neural,sennrich2015improving} is a recently proposed approach to machine translation which builds a single neural network that takes a source sentence
\begin{math} x=(x_{1},...,x_{T_x}) \end{math}
as an input and generates its translation
\begin{math} y=(y_{1},...,y_{T_y}) \end{math}
, where
\begin{math} x_{t} \end{math}
and
\begin{math} y_{t} \end{math}
are source and target symbols. Ever since the integration of attention \cite{bahdanau2014neural,cho2014properties}, NMT systems have seen remarkable improvement on translation quality. Most commonly, an attentional NMT consists of three components: (a) an encoder which computes a representation for each source sequence; (b) a decoder which generates one target symbol at a time, shown in  Eq.1 ; (c) the attention mechanism which computes a weighted global context with respect to the source and all the generated target symbols.

\begin{equation}
\log p(y|x)= \sum_{t=1}^{T_y} \log p(y_t|y_{t-1},x)
\end{equation}

Given N training sentence pairs 
\begin{math} {(x^i,y^i)\dots(x^n,y^n)\dots(x^N,y^N)} \end{math},
Maximum Likelihood Estimation (MLE) is usually accepted to optimize the model, and training objective is defined as:

\begin{align}
L_{MLE} & =-\sum_{n=1}^{N}\log  p(y^{n}|x^{n})\\
        & =-\sum_{n=1}^{N}\sum_{t=1}^{T_{y}}\log p(y_{t}^n|y_{t-1}^n,x^n)
\end{align}

Among all the encoder-decoder models, the recently proposed Transformer \cite{vaswani2017attention} architecture achieves the best translation quality so far. In this paper, we introduce the most advanced Transformer model architecture into the query translation, which greatly reduces the ambiguity of translation, and improves the quality of retrieval. 

The Transformer architecture relies on a self-attention mechanism \cite{lin2017structured} to calculate the representation of the source and target side sentences, removing all recurrent or convolutional operations found in the previous methods. Each token is directly connected to any other token in the same sentence via self-attention. The hidden state in the Transformer encoder is calculated based on all hidden states of the previous layer. The hidden state 
\begin{math} h_{t}^{i} \end{math}
in a self-attention network is calculated as in Eq.3.

\begin{equation}
h_{t}^{i}=h_{t-1}^{i}+F(self{-}attention(h_{t-1}^{i}))
\end{equation}

where \begin{math} F \end{math} represents a feed-forward network with layer normalization and ReLU as the activation function. The decoder additionally has a multi-head attention over the encoder hidden states. For more details, refer to Vaswani \cite{vaswani2017attention}.
\begin{figure}[t]
    \centering    
    \includegraphics[keepaspectratio, width=0.49\textwidth]{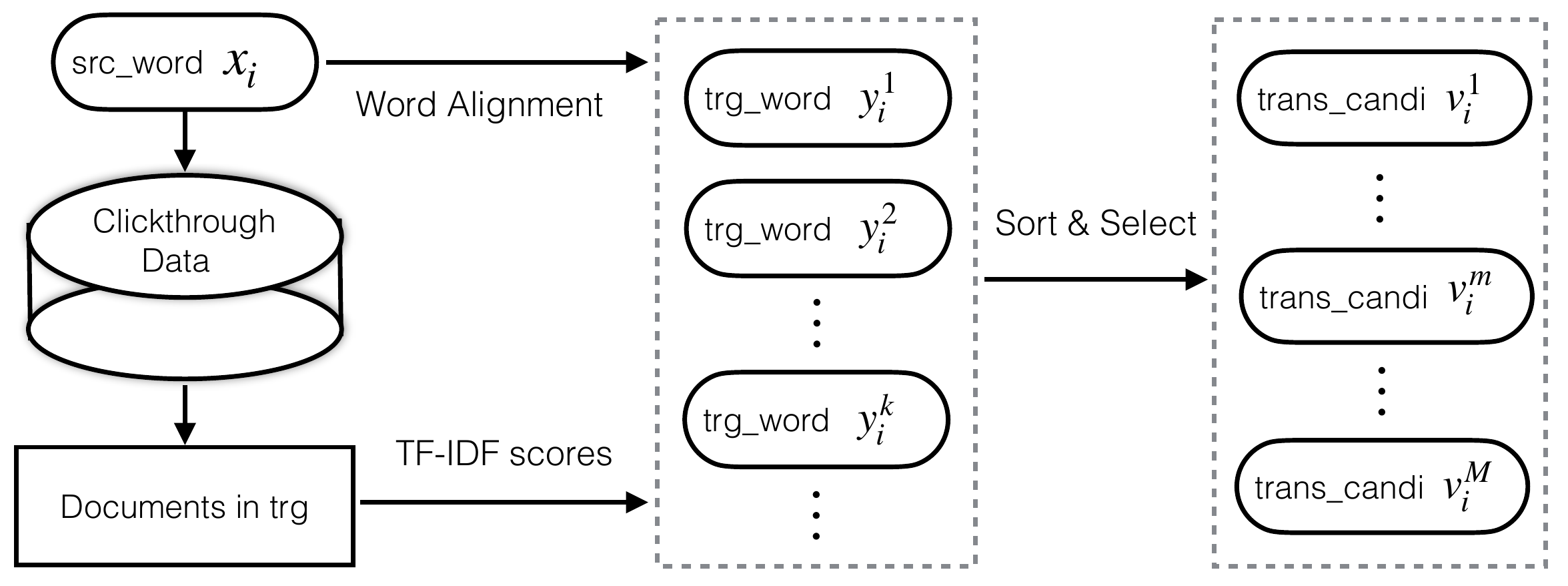}
    \caption{Illustration of the mining method for constraint translation candidates. Our approach first collects the translation candidates using word alignment, which are then sorted and filtered associated with their TF-IDF scores in the set of documents related to the given source word. }
    \label{trans_candi}
\end{figure} 

\begin{figure*}[h]
    \centering    
    \includegraphics[keepaspectratio, width=0.9\textwidth]{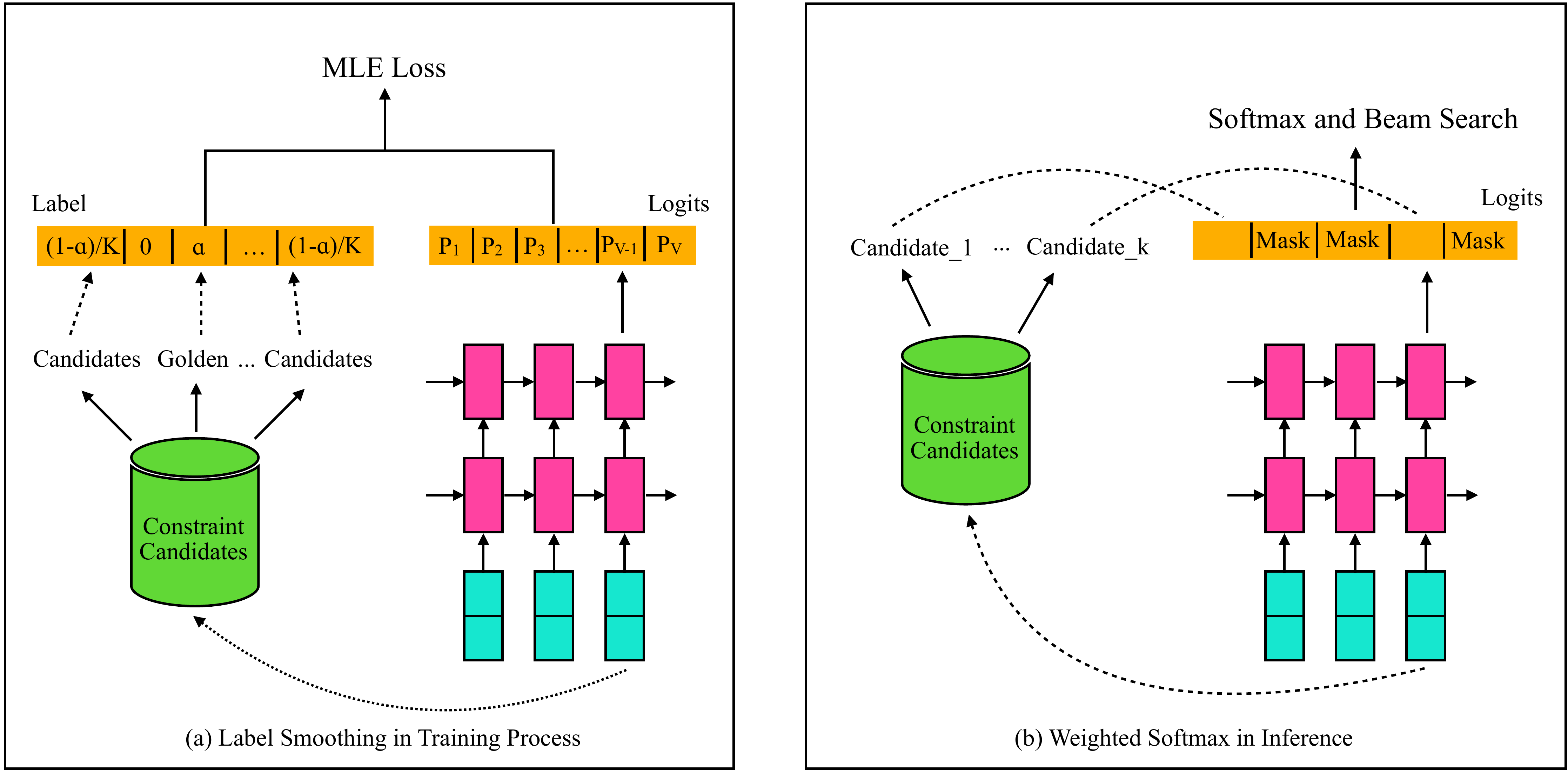}
    \caption{Illustration of training procedure with label smoothing (a) and inference procedure with weighted softmax (b).}
    \label{weighted softmax.arch}
\end{figure*}


\section{Constraint Translation Candidates}
In this section, we introduce our proposed method. The neural query translation and information retrieval is bridged with constraint translation candidates. 
This vocabulary set is mined from parallel corpus and scored according to the term frequency and inverted document frequency in search index.  Then, we employ these constraint candidates to guide NMT model to learn and generate the search-aware tokens. 
Specifically the constrained candidates will be given more weights in training stage. In inference, we will constrain the translation outputs of each query to these candidate vocabularies. 




\subsection{Mining Constraint Candidates}
Naturally, an alternative way to select the search-aware translations is to find out those important candidates that likely appear in the retrieval entries, as shown in Figure~\ref{trans_candi}. 

\noindent



\noindent
\textbf{Word Alignment}

Specifically, given a source word $x_i$ in user query $x$, we first obtain a set of its possible target tokens $C^{bi}_i$ with its translation possibility distribution in bilingual training corpus. 
This process can be achieved by a statistical word alignment tool--GIZA++\footnote{\url{https://github.com/moses-smt/giza-pp}} which is able to get alignment distribution between source and target. 
Generally, GIZA++ implements IBM Models and aligns words based on statistical models. The best alignment of one sentence pair $\widehat{\theta}^{a}$ is called Viterbi alignment:
\begin{equation}
    \widehat{\theta}^{a} = \underset{\widehat{\theta}^{a}}{argmax} p_{\widehat{\Psi }}({y}^{a}, {\theta} ^{a}|x^{a})
\end{equation} where $\Psi$ can be estimated using maximum likelihood estimation on query translation corpus:
\begin{equation}
    \widehat{\Psi} = \underset{\Psi}{argmax}\prod _{s=0}^{S}\sum _{\theta }p_{\Psi }(y^s, \theta | x^s)
\end{equation}
Here, $S$ is the size of bilingual data. $x^s$ and $y^s$ denotes the source and target sentences, respectively.  $\theta$ means weights of alignment.

\noindent
\textbf{TF-IDF}

The candidates $C^{bi}_i$ can be continually scored and filtered according to the distribution of target terms in the entries clicked by users. 
Users across the world issue multilingual queries to the search engines of a website everyday, which form large-scale cross-lingual clickthrough data.  
Intuitively, when a recalled item leads the user to click details and even make purchases, we attribute the target tokens in items satisfy the expectation of users. 
With the help of such an automatic and low cost quality estimation approach, our model can acquire high quality in-domain translation candidates derived from documents and user behaviors.

From the clickthrough data, we first extract all the documents $D_{x_i}$ that users clicked with any queries contain $x_i$. Thus, we can use TF-IDF score to identify the importance of each translation candidates in $C^{bi}_i$: 
\begin{equation}
\label{tf}
    TF-IDF_{y^{k}_i} = TF_{y^{k}_i} * IDF_{y^{k}_i}
\end{equation}

\begin{equation} \label{tf:phrasek}
    TF_{y^{k}_i} = \frac{N_{y^{k}_i}}{\sum_{m=1}^{M}{N_{y^{m}_i}}}
\end{equation} 
\begin{equation} \label{y:phrasek}
    IDF_{y^{k}_i} = log(\frac{G_{Y}}{G_{y^{k}_i} + 1})
\end{equation}
where $N_{(*)}$ indicates the frequency that the target term has appeared in $D_{x_i}$. $G_{Y}$ denotes the number of documents in $D_{x_i}$ and $G_{y_{i}^k}$ is the number of documents contain $y_{i}^k$.

Different from traditional TF-IDF which calculates scores over all the documents, our approach merely considers the documents that user clicked with a word $x_i$, thus building correlation among multi-lingual queries and words in documents.  

Finally, we can sort the items in $C^{bi}_i$, and select $M$ words which have the highest scores as constrained translation candidates $V^{candi}=\{v_1,\dots,v_m,\dots,v_M\}$. In experiments, we will explore how the size $M$ affects translation quality.



\subsection{Training with Label Smoothing}

In training process, we use the translation candidates in label smoothing. When calculating the loss of $word_{t}$, we assign a weight $\alpha$ to the golden label and $1-\alpha$ to the other constraint translation candidates related to source words equally. With this strategy, we can remove the gap between training and inference. Figure \ref{weighted softmax.arch} (a) illustrates the training procedure of our proposed method.

In training process, different from traditional MLE, we follow the equations below:

\begin{align} \label{MLE_new}
L_{MLE_{new}} &  = \alpha * L_{MLE} + ({1-\alpha}) * L_{\varphi}
\end{align}

\begin{align} \label{varphi}
    L_{\varphi} = -\sum_{i=1}^{N}\sum_{t=1}^{T_{y}}\sum_{m=1}^{M}(logp(v_{m}|{y}^n_{t-1}, {x}^n)
\end{align}

 where $M$ is the size of words picked from candidates. Contrast to conventional learning objective which merely pays attention to the ground-truth label, we offer the candidates of source words with a loss factor of $1-\alpha$, thus guiding the NMT model to generate the selected words. In our experiments, we empirically set $\alpha$ to 0.6.

\subsection{Inference with Weighted Softmax}
In NMT, the probability of a prediction is calculated by a non-linear function $softmax$. 
Given an output hidden state $h \in R^{D}$ with the hidden size being $D$, the translation probability of the $j$-th word in the vocabulary set can be formally expressed as: 
\begin{align} \label{softmax}
    p(y_j) & = \frac{exp(W_j*h + b_j)}{\sum_{k=1}^{|V|}exp(W_k*h + b_k)} 
\end{align} 
where $W \in R^{|V|\times D}$ and  $b \in R^V$ are trainable parameter matrix and bias of the vocabulary $V$, respectively.  

As seen, in the conventional approach, all the target words are considered, some of which are completely 
unrelated to the original query and the downstream search task. 
Accordingly, an alternative way to assign higher probabilities to constraint translation candidates is to locate factors in $softmax$. 
In this paper, we apply a more simple manner that normalizes the probabilities of output words in the proposed constraint space.   
\begin{align} \label{softmax}
    p(y_j) & = \frac{exp(W_j*h + b_j)}{\sum_{k=1}^{|V^{candi}|}exp(W_k*h + b_k)} 
\end{align} 
In this way, the translation model merely calculates the prediction distribution on the constraint translation candidates, thus generating more related tokens for the subsequent task. Figure \ref{weighted softmax.arch} (b) shows the basic process of translation. 


\section{Experiments}
In this section, we conducted experiments on Aliexpress Russian (Ru) to English (En) CLIR engine to evaluate the effectiveness of the proposed method.

\subsection{Data}
We train our model based on our in-house Ru-En parallel corpus which consists of about 150M general domain sentence pairs.
We build the constraint translation candidate by collecting user clickthrough data from Aliexpress e-commerce website in October 2018. 
All the Russian and English sentences are tokenized using the scripts in Moses. To avoid the problem of out of vocabulary, the sentences are processed by byte-pair encoding (BPE)  \cite{sennrich2015neural} with 32K merge operations for all the data. Accordingly, the vocabulary size of Ru and En are set to 30k. 
5K queries in search scenarios are randomly extracted and translated by human translators. We treat this dataset as the test set.

\subsection{Experimental Setting}
We build our model upon advanced Transformer model~\cite{vaswani2017attention}. Following the common setting, we set the number of layers in encoder and decoder to 6 and hidden size $D$ to 512. We employ multi-head attention with 8 attention heads and 1024 feed-forward dimensions. During training, we set the dropout rate to 0.1.
We train our model with parallelization at data
batch level with a total batch of 16,384 tokens. For Russia-English task, it takes 300K-400K steps to converge on 4 V100 GPUs.
We use Adam optimizer with $\beta_{1}$ = 0.9, $\beta_{2}$ = 0.98 and $\epsilon = 10^{-9}$. We use the same warmup and decay strategy for learning rate as Vaswani et al. \cite{vaswani2017attention}, with 8000 warmup steps.  For evaluation, we use beam search with beam size of 4 and length penalty is 0.6. All the examined models in this paper were re-implemented on the top of our in-house codes based on Tensorflow. We conduct experiments on following models:
\begin{itemize}
    \item Transformer represents the vanilla NMT baseline with the advanced self-attention-based architecture \cite{vaswani2017attention}.
    \item SMT is the phrase-based statistical system of Moses. Our constraint candidates are extracted from the phrase table generated by SMT model.
    \item +TC denotes the Transformer model enhanced with the proposed methods.
\end{itemize}

\begin{table}[t]
    \caption{Ablation study on different constraint size (M in Section 3.1). ``Training'' and ``Inference'' denote the constraint size at the training and inference time, respectively. We use BLEU as the assessment metric.}
    \label{translation-quality}
  \begin{tabular}{lrrr}
    \toprule
    Models & Training & Inference & BLEU (\%) \\
    \midrule
    \multirow{5}{*}{Transformer + {TC}}& 5 & 5 & 43.81   \\
     & 10 & 10 & \textbf{44.20}  \\
     & 10 & 30K & 44.06 \\
     & 30K & 10 & 42.13 \\     
    & 20 & 20 & 43.50  \\
    \bottomrule
  \end{tabular}
\end{table}

\begin{table}[t]
    \caption{Main results of the compared models on Ru-En query translation tasks.}
    \label{main result}
  \begin{tabular}{lr}
    \toprule
    Models   & BLEU (\%) \\
    \midrule
    \text{SMT}  & 38.04 \\
    \text{Transformer}  & 43.93 \\ 
    \text{Transformer + {TC}} & \textbf{44.20} \\
    \bottomrule
  \end{tabular}
\end{table}

\subsection{Translation Quality}
In the first series of experiments, we evaluated the impact of different constraint size on the Ru$\Rightarrow$En translation tasks. As shown in Table~\ref{translation-quality}, with the increase of the constraint size, our method consistently improves the translation quality. The result demonstrates that, a small set of constraint translation may miss some of important vocabularies,  weakening the generalization ability of the model. The larger constraint size offers a flexible manner to select predictions, thus yields better performance.  However, when the size raises to 20, the translation quality reduces. We attribute this to the fact that unrelated candidates makes error propagation from TF-IDF or word alignment, and leads to the decline of translation quality.

Moreover, we also examine the effectiveness of the candidates applied at different stage. As observed, merely constraining the vocabulary size at training time performs better than that at decoding time. We ascribe this to the open problem of exposure bias in deep learning, which is partially caused by the different data distribution between training and decoding. Applying the two strategies jointly yields highest BLEU score, indicating that the two methods are complementary to each other.
Finally, we use the best setting, i.e. 10 constraint size for both training and inference, as the default setting in subsequent experiments.

\subsection{Main Translation Results}
In this section, we evaluate the proposed approach on Ru-En query translation tasks to compare our models with baseline systems, as list in Table \ref{main result}. Our neural machine translation baseline significantly outperforms the SMT model on such kind of phrase-level text translation task, which makes the evaluation convincing. The results also confirm that the neural query translation model surpasses its SMT counterpart. As seen, the proposed model yields higher BLEU score than the strong baseline system, revealing the effectiveness of our methods to improve the translation quality of query translation.
\subsection{Retrieval Performance}
\begin{table}[t]
  \caption{Effect of the proposed methods on the downstream information retrieval task.}
  \label{Performance in Information Retrieval}
  \begin{tabular}{crr}
    \toprule
    Metrics & \text{Transformer} & \text{Transformer + TC}\\
    \midrule
    RECALL & 86.69\% & 87.02\% \\
    MAP & 29.22 & 29.47\\
    NDCG@10 & 35.55 & 35.71\\
    \bottomrule
  \end{tabular}
\end{table}

We further conduct experiments to learn whether the proposed method can improve the downstream CLIR task. We integrate the compared query translation models into our CLIR system, and  examine the retrieval accuracy of 1612 search queries in 21906 documents. The experimental results are concluded in Table \ref{Performance in Information Retrieval}. Obviously, on both of RECALL, MAP and NDCG@10 indicators, our model consistently surpass the baseline Transformer model. 
The results confirm our hypothesis that forcing the query translation model to generate search-ware tokens benefits the retrieval task. The proposed method provides an alternative way to bridge the neural query translation and information retrieval, and offers better recalled items for users. 

\subsection{Qualitative Analysis}

\begin{table}[t]
    \centering    
    \caption{Case study on translation examples output by baseline and the proposed model. ``SRC'' and ``REF'' denote the source query and its translation reference, respectively.}
    \label{Fig.Examples}
    \includegraphics[keepaspectratio, width=0.45\textwidth]{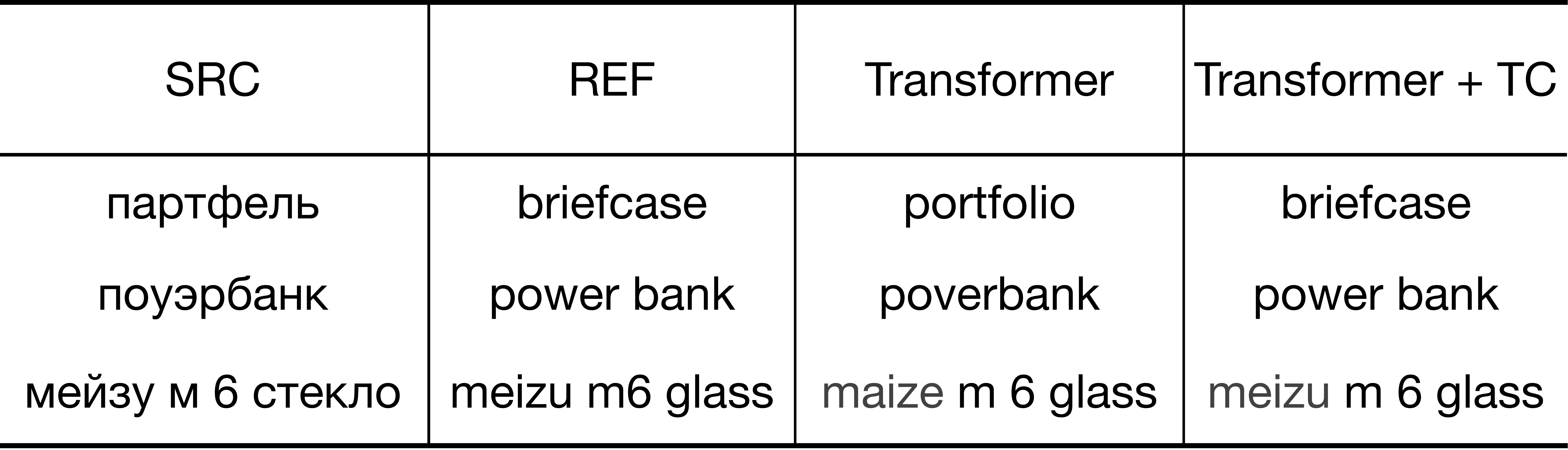}
\end{table}
In order to understand how the proposed approach exactly effects the translation and retrieval quality, we analyse the translation results in test set. As shown in Table \ref{Fig.Examples}, the case study on Russian to English translation show that, with the help of constraint translation candidates, the quality of translation is indeed improved. For example, in the baseline model which trained with general domain data, 
the brand of cell phone ``meizu'' is mistranslated. This is caused by marginal frequency of the token ``meizu'' in general training data. Thanks to the constraint translation candidates, our model correctly gets the translation. We checked our translation candidate and found that the wrong translation  ``maize'' is not appeared in the list, thus improving the translation quality.

\section{Related Work}

The correlation between MT system quality and the performance of CLIR system has been studied before. Pecina\cite{pecina2014adaptation} investigated the effect of adapting MT system to improve CLIR system. They found that the MT systems were significantly improved, but the retrieval quality of CLIR systems did not outperform the baseline system. This means that improving translation quality does not lead to improve the performance of CLIR system. Shadi\cite{Shadi16} conducted various experiments to verify that the domain of the collection that CLIR uses for retrieval and the domain of the data that was used to train MT system should be similar as much as possible for better results. 

To alleviate the mismatch between translated queries and search index, there are mainly three lines of research works. The first line is re-ranking. Re-ranking takes the alternative translations that are produced by an query translation system, re-ranks them and takes the translation that gives the best performance for CLIR in descending way. Shadi\cite{saleh2016reranking} explored a method to make use of multiple translations produced by an MT system, which are reranked using a supervised machine-learning method trained to directly optimize retrieval quality. They showed that the method could significantly improve the retrieval quality compared to a system using single translation provided by MT. The second line is optimizing translation decoder directly. Our work falls into this category. Sokolov\cite{sokolov2014learning} proposed an approach to directly optimising an translation decoder to immediately output the best translation for CLIR, which tuned translation model weights towards the retrieval objective and enabled the decoder to score the hypotheses considering the optimal weights for retrieval objective. 
The last line is multi-task learning which joint multiple tasks into training. Sarwar \cite{sarwar2019multi} proposes a multi-task learning approach to train a neural translation model with a Relevance-based Auxiliary Task (RAT) for search query translation. Their work achieves improvement over a strong NMT baseline and gets balanced and precise translations.

\section{Conclusion}

In this paper, we propose a novel approach to tackle the problem  of mismatch between neural query translation and cross-lingual information retrieval. We extract a set of constraint translation candidates that contains important words mined from search index database. 
The constraint translation candidates are incorporated into both of training and inference stages, thus instructing the translation model to learn and generate well performing target queries. 
Our model is built upon an advanced Transformer architecture and evaluated in a real-word e-Commerce search engine--Aliexpress. Experiments demonstrate that the proposed method can improve the retrieval accuracy and also maintain the translation quality.  The qualitative analysis confirms that our method exactly raises the ability of NMT to generates more suitable target queries for the real scenario of e-Commerce search.

As our approach is not limited to information retrieval tasks, it is interesting to validate the similar idea in other cross-lingual tasks that have the mismatch problem. Another promising direction is to design more powerful candidate selection techniques, e.g. calculating the distance between queries using cross-lingual pre-trained language models~\cite{devlin2019bert}. It is also interesting to combine with other techniques  \cite{li2019neuron,xu2019leveraging,zhou2020uncertainty,wan2020self} to further improve the performance of neural query translation. 

In future, we will continue to focus on how to update the constraint candidate set efficiently and use knowledge of search index to guide query translation through multi-task learning and re-ranking techniques.

\bibliographystyle{ACM-Reference-Format}
\bibliography{sample-base}


\begin{thebibliography}{31}


\ifx \showCODEN    \undefined \def \showCODEN     #1{\unskip}     \fi
\ifx \showDOI      \undefined \def \showDOI       #1{#1}\fi
\ifx \showISBNx    \undefined \def \showISBNx     #1{\unskip}     \fi
\ifx \showISBNxiii \undefined \def \showISBNxiii  #1{\unskip}     \fi
\ifx \showISSN     \undefined \def \showISSN      #1{\unskip}     \fi
\ifx \showLCCN     \undefined \def \showLCCN      #1{\unskip}     \fi
\ifx \shownote     \undefined \def \shownote      #1{#1}          \fi
\ifx \showarticletitle \undefined \def \showarticletitle #1{#1}   \fi
\ifx \showURL      \undefined \def \showURL       {\relax}        \fi
\providecommand\bibfield[2]{#2}
\providecommand\bibinfo[2]{#2}
\providecommand\natexlab[1]{#1}
\providecommand\showeprint[2][]{arXiv:#2}

\bibitem[\protect\citeauthoryear{Bahdanau, Cho, and Bengio}{Bahdanau
  et~al\mbox{.}}{2014}]%
        {bahdanau2014neural}
\bibfield{author}{\bibinfo{person}{Dzmitry Bahdanau},
  \bibinfo{person}{Kyunghyun Cho}, {and} \bibinfo{person}{Yoshua Bengio}.}
  \bibinfo{year}{2014}\natexlab{}.
\newblock \showarticletitle{Neural machine translation by jointly learning to
  align and translate}.
\newblock \bibinfo{journal}{\emph{arXiv preprint arXiv:1409.0473}}
  (\bibinfo{year}{2014}).
\newblock


\bibitem[\protect\citeauthoryear{Cho, Van~Merri{\"e}nboer, Bahdanau, and
  Bengio}{Cho et~al\mbox{.}}{2014}]%
        {cho2014properties}
\bibfield{author}{\bibinfo{person}{Kyunghyun Cho}, \bibinfo{person}{Bart
  Van~Merri{\"e}nboer}, \bibinfo{person}{Dzmitry Bahdanau}, {and}
  \bibinfo{person}{Yoshua Bengio}.} \bibinfo{year}{2014}\natexlab{}.
\newblock \showarticletitle{On the properties of neural machine translation:
  Encoder-decoder approaches}.
\newblock \bibinfo{journal}{\emph{arXiv preprint arXiv:1409.1259}}
  (\bibinfo{year}{2014}).
\newblock


\bibitem[\protect\citeauthoryear{Devlin, Chang, Lee, and Toutanova}{Devlin
  et~al\mbox{.}}{2019}]%
        {devlin2019bert}
\bibfield{author}{\bibinfo{person}{Jacob Devlin}, \bibinfo{person}{Ming-Wei
  Chang}, \bibinfo{person}{Kenton Lee}, {and} \bibinfo{person}{Kristina
  Toutanova}.} \bibinfo{year}{2019}\natexlab{}.
\newblock \showarticletitle{{BERT: Pre-training of Deep Bidirectional
  Transformers for Language Understanding}}. In
  \bibinfo{booktitle}{\emph{Proceedings of the 2019 Conference of the North
  American Chapter of the Association for Computational Linguistics: Human
  Language Technologies}}.
\newblock


\bibitem[\protect\citeauthoryear{Fujii, Utiyama, Yamamoto, and Utsuro}{Fujii
  et~al\mbox{.}}{2009}]%
        {Fujii09}
\bibfield{author}{\bibinfo{person}{Atsushi Fujii}, \bibinfo{person}{Masao
  Utiyama}, \bibinfo{person}{Mikio Yamamoto}, {and} \bibinfo{person}{Takehito
  Utsuro}.} \bibinfo{year}{2009}\natexlab{}.
\newblock \showarticletitle{Evaluating effects of machine translation accuracy
  on cross-lingual patent retrieval}.
\newblock \bibinfo{journal}{\emph{Proceedings of the 32nd international ACM
  SIGIR conference on Research and development in information retrieval}}
  (\bibinfo{year}{2009}), 674--675.
\newblock
\urldef\tempurl%
\url{https://doi.org/10.1145/1571941.1572072}
\showDOI{\tempurl}


\bibitem[\protect\citeauthoryear{Hassan, Aue, Chen, Chowdhary, Clark,
  Federmann, Huang, Junczys-Dowmunt, Lewis, Li, et~al\mbox{.}}{Hassan
  et~al\mbox{.}}{2018}]%
        {hassan2018achieving}
\bibfield{author}{\bibinfo{person}{Hany Hassan}, \bibinfo{person}{Anthony Aue},
  \bibinfo{person}{Chang Chen}, \bibinfo{person}{Vishal Chowdhary},
  \bibinfo{person}{Jonathan Clark}, \bibinfo{person}{Christian Federmann},
  \bibinfo{person}{Xuedong Huang}, \bibinfo{person}{Marcin Junczys-Dowmunt},
  \bibinfo{person}{William Lewis}, \bibinfo{person}{Mu Li}, {et~al\mbox{.}}}
  \bibinfo{year}{2018}\natexlab{}.
\newblock \showarticletitle{{Achieving Human Parity on Automatic Chinese to
  English News Translation}}.
\newblock \bibinfo{journal}{\emph{arXiv preprint arXiv:1803.05567}}
  (\bibinfo{year}{2018}).
\newblock


\bibitem[\protect\citeauthoryear{Koehn}{Koehn}{2009}]%
        {koehn2009statistical}
\bibfield{author}{\bibinfo{person}{Philipp Koehn}.}
  \bibinfo{year}{2009}\natexlab{}.
\newblock \bibinfo{booktitle}{\emph{Statistical machine translation}}.
\newblock \bibinfo{publisher}{Cambridge University Press}.
\newblock


\bibitem[\protect\citeauthoryear{Li, Wang, Yang, Shi, Lyu, and Tu}{Li
  et~al\mbox{.}}{2020}]%
        {li2019neuron}
\bibfield{author}{\bibinfo{person}{Jian Li}, \bibinfo{person}{Xing Wang},
  \bibinfo{person}{Baosong Yang}, \bibinfo{person}{Shuming Shi},
  \bibinfo{person}{Michael~R Lyu}, {and} \bibinfo{person}{Zhaopeng Tu}.}
  \bibinfo{year}{2020}\natexlab{}.
\newblock \showarticletitle{{Neuron Interaction Based Representation
  Composition for Neural Machine Translation}}. In
  \bibinfo{booktitle}{\emph{AAAI}}.
\newblock


\bibitem[\protect\citeauthoryear{Lin, Feng, Santos, Yu, Xiang, Zhou, and
  Bengio}{Lin et~al\mbox{.}}{2017}]%
        {lin2017structured}
\bibfield{author}{\bibinfo{person}{Zhouhan Lin}, \bibinfo{person}{Minwei Feng},
  \bibinfo{person}{Cicero Nogueira~dos Santos}, \bibinfo{person}{Mo Yu},
  \bibinfo{person}{Bing Xiang}, \bibinfo{person}{Bowen Zhou}, {and}
  \bibinfo{person}{Yoshua Bengio}.} \bibinfo{year}{2017}\natexlab{}.
\newblock \showarticletitle{A structured self-attentive sentence embedding}.
\newblock \bibinfo{journal}{\emph{arXiv preprint arXiv:1703.03130}}
  (\bibinfo{year}{2017}).
\newblock


\bibitem[\protect\citeauthoryear{Och and Ney}{Och and Ney}{2002}]%
        {och2002discriminative}
\bibfield{author}{\bibinfo{person}{Franz~Josef Och} {and}
  \bibinfo{person}{Hermann Ney}.} \bibinfo{year}{2002}\natexlab{}.
\newblock \showarticletitle{Discriminative training and maximum entropy models
  for statistical machine translation}. In
  \bibinfo{booktitle}{\emph{Proceedings of the 40th annual meeting on
  association for computational linguistics}}. Association for Computational
  Linguistics, \bibinfo{pages}{295--302}.
\newblock


\bibitem[\protect\citeauthoryear{Ott, Edunov, Grangier, and Auli}{Ott
  et~al\mbox{.}}{2018}]%
        {ott2018scaling}
\bibfield{author}{\bibinfo{person}{Myle Ott}, \bibinfo{person}{Sergey Edunov},
  \bibinfo{person}{David Grangier}, {and} \bibinfo{person}{Michael Auli}.}
  \bibinfo{year}{2018}\natexlab{}.
\newblock \showarticletitle{{Scaling Neural Machine Translation}}. In
  \bibinfo{booktitle}{\emph{Proceedings of the Third Conference on Machine
  Translation: Research Papers}}.
\newblock


\bibitem[\protect\citeauthoryear{Pecina, Du{\v{s}}ek, Goeuriot, Haji{\v{c}},
  Hlav{\'a}{\v{c}}ov{\'a}, Jones, Kelly, Leveling, Mare{\v{c}}ek, Nov{\'a}k,
  et~al\mbox{.}}{Pecina et~al\mbox{.}}{2014}]%
        {pecina2014adaptation}
\bibfield{author}{\bibinfo{person}{Pavel Pecina}, \bibinfo{person}{Ond{\v{r}}ej
  Du{\v{s}}ek}, \bibinfo{person}{Lorraine Goeuriot}, \bibinfo{person}{Jan
  Haji{\v{c}}}, \bibinfo{person}{Jaroslava Hlav{\'a}{\v{c}}ov{\'a}},
  \bibinfo{person}{Gareth~JF Jones}, \bibinfo{person}{Liadh Kelly},
  \bibinfo{person}{Johannes Leveling}, \bibinfo{person}{David Mare{\v{c}}ek},
  \bibinfo{person}{Michal Nov{\'a}k}, {et~al\mbox{.}}}
  \bibinfo{year}{2014}\natexlab{}.
\newblock \showarticletitle{Adaptation of machine translation for multilingual
  information retrieval in the medical domain}.
\newblock \bibinfo{journal}{\emph{Artificial intelligence in medicine}}
  \bibinfo{volume}{61}, \bibinfo{number}{3} (\bibinfo{year}{2014}),
  \bibinfo{pages}{165--185}.
\newblock


\bibitem[\protect\citeauthoryear{Popel and Bojar}{Popel and Bojar}{2018}]%
        {popel2018training}
\bibfield{author}{\bibinfo{person}{Martin Popel} {and}
  \bibinfo{person}{Ond{\v{r}}ej Bojar}.} \bibinfo{year}{2018}\natexlab{}.
\newblock \showarticletitle{{Training Tips for the Transformer Model}}.
\newblock \bibinfo{journal}{\emph{The Prague Bulletin of Mathematical
  Linguistics}} \bibinfo{volume}{110}, \bibinfo{number}{1}
  (\bibinfo{year}{2018}), \bibinfo{pages}{43--70}.
\newblock


\bibitem[\protect\citeauthoryear{Rubino}{Rubino}{2020}]%
        {rubino2020effect}
\bibfield{author}{\bibinfo{person}{Carl Rubino}.}
  \bibinfo{year}{2020}\natexlab{}.
\newblock \showarticletitle{The Effect of Linguistic Parameters in CLIR
  Performance}. In \bibinfo{booktitle}{\emph{Proceedings of the workshop on
  Cross-Language Search and Summarization of Text and Speech (CLSSTS2020)}}.
  \bibinfo{pages}{1--6}.
\newblock


\bibitem[\protect\citeauthoryear{Saleh and Pecina}{Saleh and Pecina}{2016a}]%
        {Shadi16}
\bibfield{author}{\bibinfo{person}{Shadi Saleh} {and} \bibinfo{person}{Pavel
  Pecina}.} \bibinfo{year}{2016}\natexlab{a}.
\newblock \showarticletitle{Adapting SMT Query Translation Reranker to New
  Languages in Cross-Lingual Information Retrieval.}
\newblock \bibinfo{journal}{\emph{In Medical Information Retrieval (MedIR)
  Workshop, Association for Computational Linguistics}} (\bibinfo{year}{2016}),
  1--4.
\newblock


\bibitem[\protect\citeauthoryear{Saleh and Pecina}{Saleh and Pecina}{2016b}]%
        {saleh2016reranking}
\bibfield{author}{\bibinfo{person}{Shadi Saleh} {and} \bibinfo{person}{Pavel
  Pecina}.} \bibinfo{year}{2016}\natexlab{b}.
\newblock \showarticletitle{Reranking hypotheses of machine-translated queries
  for cross-lingual information retrieval}. In
  \bibinfo{booktitle}{\emph{International Conference of the Cross-Language
  Evaluation Forum for European Languages}}. Springer, \bibinfo{pages}{54--66}.
\newblock


\bibitem[\protect\citeauthoryear{Sarwar, Bonab, and Allan}{Sarwar
  et~al\mbox{.}}{2019}]%
        {sarwar2019multi}
\bibfield{author}{\bibinfo{person}{Sheikh~Muhammad Sarwar},
  \bibinfo{person}{Hamed Bonab}, {and} \bibinfo{person}{James Allan}.}
  \bibinfo{year}{2019}\natexlab{}.
\newblock \showarticletitle{A Multi-Task Architecture on Relevance-based Neural
  Query Translation}. In \bibinfo{booktitle}{\emph{Proceedings of the 57th
  Annual Meeting of the Association for Computational Linguistics}}.
  \bibinfo{pages}{6339--6344}.
\newblock


\bibitem[\protect\citeauthoryear{Sennrich, Haddow, and Birch}{Sennrich
  et~al\mbox{.}}{2015a}]%
        {sennrich2015improving}
\bibfield{author}{\bibinfo{person}{Rico Sennrich}, \bibinfo{person}{Barry
  Haddow}, {and} \bibinfo{person}{Alexandra Birch}.}
  \bibinfo{year}{2015}\natexlab{a}.
\newblock \showarticletitle{Improving neural machine translation models with
  monolingual data}.
\newblock \bibinfo{journal}{\emph{arXiv preprint arXiv:1511.06709}}
  (\bibinfo{year}{2015}).
\newblock


\bibitem[\protect\citeauthoryear{Sennrich, Haddow, and Birch}{Sennrich
  et~al\mbox{.}}{2015b}]%
        {sennrich2015neural}
\bibfield{author}{\bibinfo{person}{Rico Sennrich}, \bibinfo{person}{Barry
  Haddow}, {and} \bibinfo{person}{Alexandra Birch}.}
  \bibinfo{year}{2015}\natexlab{b}.
\newblock \showarticletitle{Neural machine translation of rare words with
  subword units}.
\newblock \bibinfo{journal}{\emph{arXiv preprint arXiv:1508.07909}}
  (\bibinfo{year}{2015}).
\newblock


\bibitem[\protect\citeauthoryear{Sharma and Mittal}{Sharma and Mittal}{2019}]%
        {sharma2019refined}
\bibfield{author}{\bibinfo{person}{Vijay Sharma} {and} \bibinfo{person}{Namita
  Mittal}.} \bibinfo{year}{2019}\natexlab{}.
\newblock \showarticletitle{Refined stop-words and morphological variants
  solutions applied to Hindi-English cross-lingual information retrieval}.
\newblock \bibinfo{journal}{\emph{Journal of Intelligent \& Fuzzy Systems}}
  \bibinfo{volume}{36}, \bibinfo{number}{3} (\bibinfo{year}{2019}),
  \bibinfo{pages}{2219--2227}.
\newblock


\bibitem[\protect\citeauthoryear{Sokolov, Hieber, and Riezler}{Sokolov
  et~al\mbox{.}}{2014}]%
        {sokolov2014learning}
\bibfield{author}{\bibinfo{person}{Artem Sokolov}, \bibinfo{person}{Felix
  Hieber}, {and} \bibinfo{person}{Stefan Riezler}.}
  \bibinfo{year}{2014}\natexlab{}.
\newblock \showarticletitle{Learning to translate queries for CLIR}. In
  \bibinfo{booktitle}{\emph{Proceedings of the 37th international ACM SIGIR
  conference on Research \& development in information retrieval}}. ACM,
  \bibinfo{pages}{1179--1182}.
\newblock


\bibitem[\protect\citeauthoryear{Vaswani, Shazeer, Parmar, Uszkoreit, Jones,
  Gomez, Kaiser, and Polosukhin}{Vaswani et~al\mbox{.}}{2017}]%
        {vaswani2017attention}
\bibfield{author}{\bibinfo{person}{Ashish Vaswani}, \bibinfo{person}{Noam
  Shazeer}, \bibinfo{person}{Niki Parmar}, \bibinfo{person}{Jakob Uszkoreit},
  \bibinfo{person}{Llion Jones}, \bibinfo{person}{Aidan~N Gomez},
  \bibinfo{person}{{\L}ukasz Kaiser}, {and} \bibinfo{person}{Illia
  Polosukhin}.} \bibinfo{year}{2017}\natexlab{}.
\newblock \showarticletitle{Attention is all you need}. In
  \bibinfo{booktitle}{\emph{Advances in Neural Information Processing
  Systems}}. \bibinfo{pages}{5998--6008}.
\newblock


\bibitem[\protect\citeauthoryear{Wan, Yang, Wong, Chao, Du, and Ao}{Wan
  et~al\mbox{.}}{2020a}]%
        {wan2020AAAI}
\bibfield{author}{\bibinfo{person}{Yu Wan}, \bibinfo{person}{Baosong Yang},
  \bibinfo{person}{Derek~F. Wong}, \bibinfo{person}{Lidia~S. Chao},
  \bibinfo{person}{Haihua Du}, {and} \bibinfo{person}{Ben~CH Ao}.}
  \bibinfo{year}{2020}\natexlab{a}.
\newblock \showarticletitle{{Unsupervised Neural Dialect Translation with
  Commonality and Diversity Modeling}}. In \bibinfo{booktitle}{\emph{AAAI}}.
\newblock


\bibitem[\protect\citeauthoryear{Wan, Yang, Wong, Zhou, Chao, Zhang, and
  Chen}{Wan et~al\mbox{.}}{2020b}]%
        {wan2020self}
\bibfield{author}{\bibinfo{person}{Yu Wan}, \bibinfo{person}{Baosong Yang},
  \bibinfo{person}{Derek~F. Wong}, \bibinfo{person}{Yikai Zhou},
  \bibinfo{person}{Lidia~S. Chao}, \bibinfo{person}{Haibo Zhang}, {and}
  \bibinfo{person}{Boxing Chen}.} \bibinfo{year}{2020}\natexlab{b}.
\newblock \showarticletitle{{Self-Paced Learning for Neural Madchine
  Translation}}. In \bibinfo{booktitle}{\emph{EMNLP}}.
\newblock


\bibitem[\protect\citeauthoryear{Wu and He}{Wu and He}{2010}]%
        {wu2010study}
\bibfield{author}{\bibinfo{person}{Dan Wu} {and} \bibinfo{person}{Daqing He}.}
  \bibinfo{year}{2010}\natexlab{}.
\newblock \showarticletitle{A study of query translation using google machine
  translation system}. In \bibinfo{booktitle}{\emph{Computational Intelligence
  and Software Engineering (CiSE), 2010 International Conference on}}. IEEE,
  \bibinfo{pages}{1--4}.
\newblock


\bibitem[\protect\citeauthoryear{Xu, Wong, Yang, Zhang, and Chao}{Xu
  et~al\mbox{.}}{2019}]%
        {xu2019leveraging}
\bibfield{author}{\bibinfo{person}{Mingzhou Xu}, \bibinfo{person}{Derek~F.
  Wong}, \bibinfo{person}{Baosong Yang}, \bibinfo{person}{Yue Zhang}, {and}
  \bibinfo{person}{Lidia~S. Chao}.} \bibinfo{year}{2019}\natexlab{}.
\newblock \showarticletitle{{Leveraging Local and Global Patterns for
  Self-Attention Networks}}. In \bibinfo{booktitle}{\emph{ACL}}.
\newblock


\bibitem[\protect\citeauthoryear{Yang, Wong, Chao, and Zhang}{Yang
  et~al\mbox{.}}{2020}]%
        {Yang2020improve}
\bibfield{author}{\bibinfo{person}{Baosong Yang}, \bibinfo{person}{Derek~F.
  Wong}, \bibinfo{person}{Lidia~S. Chao}, {and} \bibinfo{person}{Min Zhang}.}
  \bibinfo{year}{2020}\natexlab{}.
\newblock \showarticletitle{{Improving Tree-based Neural Machine Translation
  with Dynamic Lexicalized Dependency Encoding}}.
\newblock \bibinfo{journal}{\emph{Knowledge-Based System}}
  \bibinfo{volume}{188} (\bibinfo{year}{2020}).
\newblock


\bibitem[\protect\citeauthoryear{Yao, Yang, Zhang, Chen, and Luo}{Yao
  et~al\mbox{.}}{2020a}]%
        {yao2020domain}
\bibfield{author}{\bibinfo{person}{Liang Yao}, \bibinfo{person}{Baosong Yang},
  \bibinfo{person}{haibo Zhang}, \bibinfo{person}{Boxing Chen}, {and}
  \bibinfo{person}{Weihua Luo}.} \bibinfo{year}{2020}\natexlab{a}.
\newblock \showarticletitle{Domain Transfer based Data Augmentation for Neural
  Query Translation}. In \bibinfo{booktitle}{\emph{COLING}}.
\newblock


\bibitem[\protect\citeauthoryear{Yao, Yang, Zhang, Luo, and Chen}{Yao
  et~al\mbox{.}}{2020b}]%
        {yao2020exploiting}
\bibfield{author}{\bibinfo{person}{Liang Yao}, \bibinfo{person}{Baosong Yang},
  \bibinfo{person}{haibo Zhang}, \bibinfo{person}{Weihua Luo}, {and}
  \bibinfo{person}{Boxing Chen}.} \bibinfo{year}{2020}\natexlab{b}.
\newblock \showarticletitle{Exploiting Neural Query Translation into Cross
  Lingual Information Retrieval}. In \bibinfo{booktitle}{\emph{SIGIR eCom}}.
\newblock


\bibitem[\protect\citeauthoryear{Yarmohammadi, Ma, Hisamoto, Rahman, Wang, Xu,
  Povey, Koehn, and Duh}{Yarmohammadi et~al\mbox{.}}{2019}]%
        {yarmohammadi2019robust}
\bibfield{author}{\bibinfo{person}{Mahsa Yarmohammadi}, \bibinfo{person}{Xutai
  Ma}, \bibinfo{person}{Sorami Hisamoto}, \bibinfo{person}{Muhammad Rahman},
  \bibinfo{person}{Yiming Wang}, \bibinfo{person}{Hainan Xu},
  \bibinfo{person}{Daniel Povey}, \bibinfo{person}{Philipp Koehn}, {and}
  \bibinfo{person}{Kevin Duh}.} \bibinfo{year}{2019}\natexlab{}.
\newblock \showarticletitle{Robust Document Representations for Cross-Lingual
  Information Retrieval in Low-Resource Settings}. In
  \bibinfo{booktitle}{\emph{Proceedings of Machine Translation Summit XVII
  Volume 1: Research Track}}. \bibinfo{pages}{12--20}.
\newblock


\bibitem[\protect\citeauthoryear{Zhou, Truran, Brailsford, Wade, and
  Ashman}{Zhou et~al\mbox{.}}{2012}]%
        {zhou2012translation}
\bibfield{author}{\bibinfo{person}{Dong Zhou}, \bibinfo{person}{Mark Truran},
  \bibinfo{person}{Tim Brailsford}, \bibinfo{person}{Vincent Wade}, {and}
  \bibinfo{person}{Helen Ashman}.} \bibinfo{year}{2012}\natexlab{}.
\newblock \showarticletitle{Translation techniques in cross-language
  information retrieval}.
\newblock \bibinfo{journal}{\emph{ACM Computing Surveys (CSUR)}}
  \bibinfo{volume}{45}, \bibinfo{number}{1} (\bibinfo{year}{2012}),
  \bibinfo{pages}{1--44}.
\newblock


\bibitem[\protect\citeauthoryear{Zhou, Yang, Wong, Wan, and Chao}{Zhou
  et~al\mbox{.}}{2020}]%
        {zhou2020uncertainty}
\bibfield{author}{\bibinfo{person}{Yikai Zhou}, \bibinfo{person}{Baosong Yang},
  \bibinfo{person}{Derek~F. Wong}, \bibinfo{person}{Yu Wan}, {and}
  \bibinfo{person}{Lidia~S. Chao}.} \bibinfo{year}{2020}\natexlab{}.
\newblock \showarticletitle{{Uncertainty-Aware Curriculum Learning for Neural
  Machine Translation}}. In \bibinfo{booktitle}{\emph{ACL}}.
\newblock


\end{thebibliography}

\end{document}